\documentclass[]{article}
\usepackage{aaai24}  
\usepackage{times}  
\usepackage{helvet}  
\usepackage{courier}  
\usepackage[hyphens]{url}  
\usepackage{graphicx} 
\usepackage{natbib}  
\usepackage{caption} 
\usepackage{algorithm}
\usepackage{algorithmic}
\usepackage{url}            
\usepackage{booktabs}       
\usepackage{amsfonts}       
\usepackage{nicefrac}       
\usepackage{microtype}      
\usepackage{xcolor}         
\usepackage{graphicx} 
\usepackage{amsmath}
\usepackage{amsthm}
\usepackage{subfig}
\usepackage{multirow,diagbox,enumitem}
\usepackage{mathrsfs}
\usepackage{tikz}
\usepackage{subfig}
\usepackage{bm}
\usepackage{amssymb}

\newcommand{\vpara}[1]{\vspace{0.05in}\noindent\textbf{#1 }}

\usepackage{newfloat}
\usepackage{listings}
\DeclareCaptionStyle{ruled}{labelfont=normalfont,labelsep=colon,strut=off} 
\floatstyle{ruled}
\newfloat{listing}{tb}{lst}{}
\floatname{listing}{Listing}
\title{Journey to the Center of the Knowledge Neurons: Discoveries of Language-Independent Knowledge Neurons and Degenerate Knowledge Neurons}
\author{
    Yuheng Chen\textsuperscript{\rm 1,\rm 2 \thanks{These authors contributed equally to this work.}}
    Pengfei Cao\textsuperscript{\rm 1,\rm 2 *},
    Yubo Chen\textsuperscript{\rm 1,\rm 2 \thanks{Corresponding author.}},
    Kang Liu\textsuperscript{\rm 1,\rm 2},
    Jun Zhao\textsuperscript{\rm 1,\rm 2}
}
\affiliations{
    \textsuperscript{\rm 1} The Laboratory of Cognition and Decision Intelligence for Complex Systems, \\
    Institute of Automation, Chinese Academy of Sciences, Beijing, China\\
    \textsuperscript{\rm 2} School of Artificial Intelligence, University of Chinese Academy of Sciences, Beijing, China\\

    chenyuheng22@ia.ac.cn, \{pengfei.cao, yubo.chen, kliu, jzhao\}@nlpr.ia.ac.cn
}
\usepackage{bibentry}

\begin{document}

\maketitle

\begin{abstract}
Pre-trained language models (PLMs) contain vast amounts of factual knowledge, but how the knowledge is stored in the parameters remains unclear. This paper delves into the complex task of understanding how factual knowledge is stored in multilingual PLMs, and introduces the Architecture-adapted Multilingual Integrated Gradients method, which successfully localizes knowledge neurons more precisely compared to current methods, and is more universal across various architectures and languages. Moreover, we conduct an in-depth exploration of knowledge neurons, leading to the following two important discoveries: (1) The discovery of Language-Independent Knowledge Neurons, which store factual knowledge in a form that transcends language. We design cross-lingual knowledge editing experiments, demonstrating that the PLMs can accomplish this task based on language-independent neurons; (2) The discovery of Degenerate Knowledge Neurons, a novel type of neuron showing that different knowledge neurons can store the same fact. Its property of functional overlap endows the PLMs with a robust mastery of factual knowledge. We design fact-checking experiments, proving that the degenerate knowledge neurons can help the PLMs to detect wrong facts. Experiments corroborate these findings, shedding light on the mechanisms of factual knowledge storage in multilingual PLMs, and contribute valuable insights to the field. The code is available at https://github.com/heng840/AMIG.
\end{abstract}

\section{Introduction}
Pre-trained language models (PLMs) \cite{mbert,gpt2,mgpt,openai2023gpt4,touvron2023llama} have revolutionized the field of natural language processing, due to their exceptional performance across a broad spectrum of tasks. 
These models, trained on extensive corpora such as Wikipedia, are widely believed to encapsulate vast amounts of factual knowledge~\cite{knowledge-base,knowledge_know_fact}, but how the knowledge is stored in the parameters remains unclear \cite{kandpal2023large}.
Investigating knowledge storage mechanisms will facilitate deeper comprehension and mastery of knowledge in PLMs \cite{zhen2022survey,zhao2023survey}. In this paper, we conduct an in-depth study on the Knowledge Localization task \cite{localization,kl}, which seeks to determine the storage location of specific factual knowledge in the model parameters, where such parameters are named \textit{Knowledge Neurons} \cite{dai2022kn}.

\begin{figure}
  \centering
  \subfloat[Language-Independent Knowledge Neurons: Acquisition process and functionality.]{
    \includegraphics[width=0.95\linewidth]{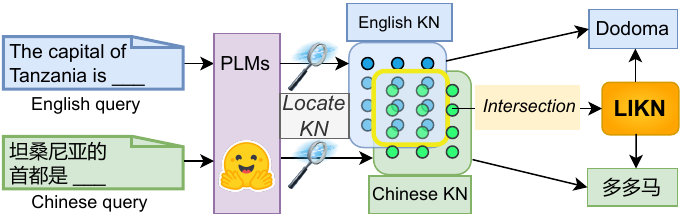}
    \label{fig-intro1}
  }
  \\
  \subfloat[Degenerate Knowledge Neurons: Acquisition process and functionality. ``ON" indicates the PLMs must activate at least one corresponding degenerate knowledge neuron.]{
    \includegraphics[width=0.95\linewidth]{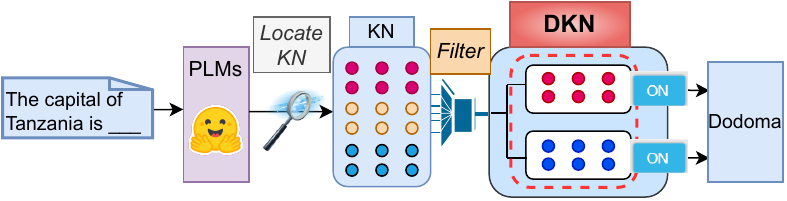}
    \label{fig:intro2}
  }
  \caption{Explanation of Language-Independent Knowledge Neurons (LIKN) and Degenerate Knowledge Neurons (DKN). KN denotes knowledge neurons.}
  \label{fig-intro}
\end{figure}

Recently, several established approaches strive to elucidate the knowledge storage mechanism in PLMs. 
One strategy is the gradient-based method \cite{ancona2019gradient}, which assesses the contribution of each neuron by calculating its attribution score using integrated gradients. Another is the causal-inspired method \cite{cao2023life}, which employs a tracing algorithm to follow causal influences across model layers. Despite successful efforts in the knowledge localization task, these methods still face two major challenges: 
(1) \textit{Lack of Universal Method for Different PLM Architectures}:
Factual knowledge is observed to emerge in all kinds of PLM architectures, including auto-encoding models (e.g., BERT) \cite{mbert} and auto-regressive models (e.g., GPT) \cite{mgpt}. However, while some methods are suited for auto-encoding models and perform poorly with auto-regressive models \cite{meng2022locating}, others are designed specifically for auto-regressive models and are not well-adapted to auto-encoding models\cite{causal-inspired}, leaving a gap in a universal approach that performs well across both PLM architectures.
(2) \textit{Lack of Exploration in Multiple Languages}:
Substantial knowledge is independent of language, and current LLMs support multilingualism. However, existing methods, with their focus solely on English datasets, may fail to provide comprehensive insights into the knowledge storage mechanism across different languages, limiting the ability to draw multilingual conclusions.

In order to localize knowledge neurons more precisely, we follow the gradient-based method and propose a novel knowledge localization method, termed Architecture-adapted Multilingual Integrated Gradients (AMIG).
Firstly, for the lack of universal method in different PLM architectures, we design an architecture adaptation technique, making the baseline vectors in the integrated gradients algorithm \cite{rigorousIG} universally compatible across different PLM architectures. Secondly, for the lack of exploration in multiple languages, we introduce a multilingual threshold adjustment technique, adjusting the thresholds in the integrated gradient calculations for different languages. Experimental results on multilingual datasets demonstrate that our method can localize the knowledge neurons more precisely compared to previous state-of-the-art models. In addition, we also conduct an in-depth exploration of knowledge neurons, leading to the following two important discoveries.

\textbf{Language-Independent Knowledge Neurons}: We discover a new type of neuron in multilingual PLMs that is capable of storing factual knowledge across languages.
We name them \textit{Language-Independent Knowledge Neurons}, since their existence transcends the boundaries of specific languages.

As illustrated in Figure \ref{fig-intro1}, these neurons are obtained by intersecting knowledge neurons derived from different languages, encapsulating knowledge representations that are consistent across multiple languages. Language-independent knowledge neurons can help cross-lingual knowledge editing tasks: a single edit to certain knowledge can simultaneously affect the corresponding knowledge in all languages. For example, if we edit the language-independent neuron corresponding to the fact $\langle  \textit{Tanzania}, \textit{Capital}, \textit{Dar es Salaam} \rangle$ to $\langle \textit{Tanzania}, \textit{Capital}, \textit{Dodoma} \rangle$, this fact will be changed correspondingly in all languages. 
We design experiments to verify the role of language-independent knowledge neurons. Compared with existing cross-lingual knowledge editing models, the editing performance of our method is superior. This experiment demonstrates the potential of our method in cross-lingual knowledge editing applications.

\textbf{Degenerate Knowledge Neurons}: We discover an interesting phenomenon, corresponding to a completely new type of neurons. Given a fact and its corresponding knowledge neurons, some subsets of knowledge neurons exhibit unique properties. Even if some elements in this subset are suppressed, the model can still express the fact correctly; however, if all elements in the subset are suppressed, the model can no longer express the fact correctly. This phenomenon demonstrates that some knowledge neurons store the same factual knowledge, and the model needs to activate at least one of the neurons to express the facts correctly. It is very similar to the ``degenerate" phenomenon in biological systems \cite{degenerate_biological,mason2015degeneracy}, so we name this type of neuron \textit{Degenerate Knowledge Neurons}. 
Unlike redundancy, degenerate knowledge neurons cannot simply be deleted because they only partially overlap. A degenerate knowledge neuron may store multiple pieces of factual knowledge, the deletion of it has no effect on specific knowledge but may affect other knowledge.

Figure \ref{fig:intro2} illustrates the acquisition process of degenerate knowledge neurons. In detail, we first localize the knowledge neurons, then aggregate and filter them to obtain degenerate knowledge neurons. For the query ``\textit{The capital of Tanzania is \underline{\hspace{0.4cm}}}", the PLM must activate at least one corresponding degenerate knowledge neuron to predict the correct fact \textit{Dodoma}.
Intuitively, the property of functional overlap in degenerate knowledge neurons endows the PLMs with a robust understanding of factual knowledge, ensuring that its mastery of facts remains stable and less prone to errors. 
Inspired by this, we design an experiment to use degenerate knowledge neurons for fact-checking. Our experiment demonstrates that the degenerate knowledge neurons can help the PLMs to detect wrong facts, thus illustrating that their presence enhances the PLMs' stable mastery of factual knowledge.

Overall, the main contributions are summarized as follows:

(1) We propose a novel knowledge localization method named architecture-adapted multilingual integrated gradients, which can effectively address the two challenges of traditional methods: the lack of a universal method for different PLM architectures and the lack of exploration in multiple languages, thus achieving more precise localization of knowledge neurons.

(2) We discover language-independent knowledge neurons, which store factual knowledge in a form that transcends language barriers. Experimental results demonstrate that they are beneficial for the cross-lingual knowledge editing task.

(3) We discover degenerate knowledge neurons, a new type of neuron that possesses properties of functional overlap, making the model's mastery of factual knowledge more robust. Experiments prove that they can help detect incorrect facts.

\section{Methodology}
\begin{figure*}[t]
\centering
\includegraphics[width=0.7\linewidth]{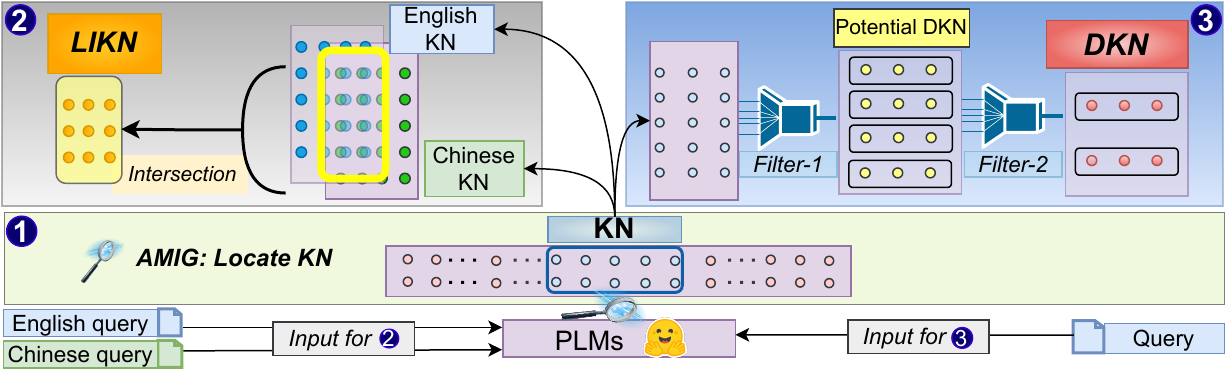}
\caption{Overall Algorithm Flow, describing (1) our architecture-adapted multilingual integrated gradients (AMIG) method for locating knowledge neurons (KN), (2) the process of detecting language-independent knowledge neurons (LIKN), and (3) the process of detecting degenerate knowledge neurons (DKN).}
\label{fig:overall}
\end{figure*}

Figure \ref{fig:overall} schematically visualizes our proposed framework. It consists of three main modules, including knowledge neuron localization (module 1), language-independent knowledge neuron detection (module 2), and degenerate knowledge neuron detection (module 3). We illustrate each module in detail.
\subsection{Knowledge Neuron Localization}
Module 1 of Figure \ref{fig:overall} showcases the knowledge localization module, which aims to pinpoint the exact locations of the knowledge neurons within a PLM. Using the fill-in-the-blank cloze task \cite{fill-in-the-blank}, we evaluate the understanding of a PLM of specific facts. For example, given a fact $\langle \textit{Tanzania}, \textit{Capital}, \textit{Dodoma} \rangle$ with corresponding query ``\textit{The capital of Tanzania is \underline{\hspace{0.4cm}}}", \citeauthor{fill-in-the-blank}\shortcite{fill-in-the-blank} describe that a model knows a fact if it can predict the correct answer. In this study, we extend this analysis by introducing the Architecture Adapted Multilingual Integrated Gradients method to localize the neurons responsible for processing factual information specifically.

Mathematically, given a query $q$, the probability of the correct answer predicted by a PLM can be defined as:
{\small \begin{align}
\label{eq:1}
      \operatorname{F}(\hat{w}^{(l)}_{j}) = p(y^* | q, w^{(l)}_{j}=\hat{w}^ {(l)}_{j}),
\end{align}}where $y^*$ is the correct answer, ${w}^{(l)}_{j}$ is the $j$-th neuron of $l$-th layer, and $\hat{w}^ {(l)}_{j}$ is a value that ${w}^ {(l)}_{j}$ is assigned to. To compute the attribution score for each neuron, we use integrated gradients\cite{ig}. Consider a neuron ${w}^ {(l)}_{j}$, we can calculate its attribution score:
{\small 
\begin{align}
    \label{eqution:attribute}
     \operatorname{Attr}({w}^{(l)}_{j}) = (\overline{w}^{(l)}_{j} - {w'}^{(l)}_{j}) \int_{0}^{1} \frac{\partial \operatorname{F}({w'}^{(l)}_{j} + \alpha(\overline{w}^{(l)}_{j} - {w'}^{(l)}_{j}))}{\partial {w}^{(l)}_{j}}  \, d\alpha,
\end{align}}where $\overline{w}^{(l)}_{j}$ is the value of ${w}^{(l)}_{j}$, ${w'}^{(l)}_{j}$ is the baseline vector of ${w}^{(l)}_{j}$, and $\frac{\partial \operatorname{F}({w'}^{(l)}_{j} + \alpha({w}^{(l)}_{j} - {w'}^{(l)}_{j}))}{\partial {w}^{(l)}_{j}}$ calculates the gradient. As $\alpha$ changes from 0 to 1, $({w'}^{(l)}_{j} + \alpha({w}^{(l)}_{j} - {w'}^{(l)}_{j}))$ changes from ${w'}^{(l)}_{j}$ to ${w}^{(l)}_{j}$, so the $\operatorname{Attr}({w}^{(l)}_{j})$ can accumulates the probability changes caused by the change of ${w}^{(l)}_{j}$ through integrating the gradients.
The ideal baseline vector ${w'}^{(l)}_{j}$, typically approximated by a zero vector \cite{transparent}, lacks consideration for diverse PLM architectures, resulting in sub-optimal performance. We address this by introducing an architecture adaptation technique to compute baseline vectors suitable for different PLM architectures.

First, in order to minimize the information content in the baseline vectors, we follow the method of \citeauthor{enguehard2023sequential}\shortcite{enguehard2023sequential}, dividing the input query $q$ into $m$ words, and then feeding each word separately into the PLM to calculate the activation score for the neurons corresponding to each word $q_i$.
Then, we meticulously design the baseline vectors for different PLM architectures. Let the baseline sentence corresponding to $q_i$ be $q'_i$, and $q'_i$ contains $m$ words, with a length consistent with $q$, denoted as $q'_i=(q'_{i1} \ldots q'_{ik} \ldots q'_{im})$, where:
{\small
\begin{align}
    q'_{ik} =
    \begin{cases}
    \langle \text{mask}\rangle  & \text{if } k = i \text{ (for auto-encoding models)} \\
    \langle \text{eos}\rangle  & \text{if } k = i \text{ (for auto-regressive models)} \\
    q_k & \text{otherwise}
    \end{cases},
\end{align}
}where $\langle \text{mask}\rangle$ is used for masking auto-encoding models, $\langle \text{eos}\rangle$ stands for ``end of sequence" in auto-regressive models, and $q_k$ is the $k$-th word of the query. In this design, the $i$-th neuron in the $l$-th layer, represented by $w_j^{(l)}$, corresponds to $q_i$,  and its associated baseline vector ${w'}^{(l)}_{j}$ corresponds to $q'_i$. We can then calculate the attribution score $Attr_i(w_j^{(l)})$ for each neuron when $q_i$ is used as input, according to Equation \eqref{eqution:attribute}.
To calculate the integral, we use the Riemann approximation:
{\small 
\begin{align}
    {Attr_i(w_j^l)} \approx \frac{\overline{w}^{(l)}_{j}}{N} \sum_{k=1}^{N} \frac{ \partial F({w'}^{(l)}_{j} + \frac{k}{N} \times (\overline{w}^{(l)}_{j} - {w'}^{(l)}_{j})}{\partial {w}^{(l)}_{j}},
\end{align}}where $N$ is the number of approximation steps. The attribution for each word $q_i$ is then summed and normalized, leading to the final attribution score for the query:
{\small \begin{align}
Attr(w_j^l) = \frac{\sum_{i=1}^{m} Attr_i(w_j^l)}{\sum_{j=1}^{n} \sum_{i=1}^{m} Attr_i(w_j^l)},
\end{align}}where $n$ is the number of neurons in the $l$-th layer. Finally, we can find the neurons with attribution scores greater than the threshold $\tau$, and consider them as knowledge neurons, denote as $\mathcal{N}$.

\subsection{Language-Independent Knowledge Neuron Detection}
\vpara{Explanation} Many PLMs support multilingualism, and a significant portion of factual knowledge within these models is language-independent \cite{m-language-edit, wang-etal-2020-negative}. This necessity has become increasingly important in exploring the storage mechanism of factual knowledge in multilingual PLMs. We define neurons that store factual knowledge common to multiple languages as \textit{Language-Independent Knowledge Neurons}, denoted as $\mathcal{L}$. To identify these type of knowledge neurons, we devise a detection algorithm that is illustrated as follows.

\vpara{Algorithm} As shown in the module 2 of Figure \ref{fig:overall}, given factual triples in $K$ languages with identical semantics, let the corresponding queries be denoted by $q^k$ for $k=1, 2, \ldots, K$. For each query, we use knowledge neuron localization module to obtain the corresponding knowledge neurons, where the attribution score of neuron $w_i^{(l)}$ is recorded as $Attr_k(w_i^{(l)})$. The sensitivity of the multilingual PLMs to different languages varies, resulting in significant differences in attribution scores for queries in different languages. Therefore, it is difficult to obtain knowledge neurons for all languages by setting a unified threshold. To solve this problem, we design a multilingual threshold adjustment technique. We set different scaling factors $\tau_k$ for different languages, and record the maximum attribution score of the neurons corresponding to query $q_k$, and then determine the dynamic threshold:
{\small\begin{align}
    T_k = \max_{i, l} Attr_k(w_i^{(l)}) \times \tau_k,
\end{align}}Then, we identify knowledge neurons \(\mathcal{N}_k\) for the \(k\)-th language using threshold filtering as follows:
{\small\begin{align}
    \mathcal{N}_k = \left\{ w_i^{(l)} \mid  Attr_k(w_i^{(l)}) > T_k,  \forall i, l \right\},
\end{align}}Finally, we compute the intersection of the knowledge neurons across all languages:
{\small\begin{align}
    \mathcal{L} = \bigcap_{k=1}^{K} \mathcal{N}_k,
\end{align} }where $\mathcal{L}$ represents the language-independent knowledge neurons, encoding factual knowledge consistent across all considered languages. Through the aforementioned algorithm, we can ultimately obtain them.

\subsection{Degenerate Knowledge Neuron Detection}
\vpara{Explanation}
By conducting in-depth analysis, we identify an intriguing phenomenon: distinct sets of neurons are responsible for storing identical factual knowledge. For example, for a specific fact denoted as \(\langle h,r,t\rangle \), suppose we localize 10 knowledge neurons labeled as \(N = \{1, 2, \ldots, 10\}\). If we suppress the neurons of sets \(A = \{1, 2\}\) or \(B = \{3, 4, 5\}\), both subsets of $N$, we observe no significant decrease in prediction probability. Conversely, suppression of the neurons of these two sets simultaneous (i.e., \(A \cup B\)) leads to a substantial loss of prediction probability. This suggests that both sets \(A\) and \(B\) house the same factual knowledge, at least one must be active for the model to accurately comprehend the fact. Furthermore, these two sets of neurons are not mutually redundant. That is to say, besides the fact \(\langle h,r,t\rangle \), \(A\) may also store the fact \(\langle h_1, r_1, t_1\rangle \), while \(B\) may store \(\langle h_2, r_2, t_2\rangle \), thus playing additional roles in PLMs.
Given the resemblance of this behavior to the \textit{degenerate} phenomenon in biological neural networks \cite{degenerate_biological,mason2015degeneracy}, we coin the term \textit{Degenerate Knowledge Neurons} for these neurons. This concept is introduced in detail next.

\vpara{Algorithm}
Formally, let $\mathcal{N}=\{n_1, \ldots, n_k\}$ be the set of all localized knowledge neurons\footnote{This can be further generalized, where each $n_k$ is itself a set, making $\mathcal{N}$ a set of sets.}, we define degenerate knowledge neurons as $\mathcal{D} = \{d^{\mathcal{D}}_1, \ldots, d^{\mathcal{D}}_m \}$, where each $d^{\mathcal{D}}_i=\{n_{i1}, \ldots, n_{iv}\}$ contains $v$ knowledge neurons, and satisfies the following conditions:
{\small \begin{align}
Prob(\mathcal{N}) - Prob(\mathcal{N}\setminus P_{\text{s}}(n_i)) &\leq T_{\text{low}}, \forall P_{\text{s}}(n_i), \label{equation-1}\\
Prob(\mathcal{N}) - Prob(\mathcal{N}\setminus \bigcup_{j=1}^v n_{ij}) &> T_{\text{high}} \label{equation-2},
\end{align}}where $P_{\text{s}}(n_i)$ is a proper subset of the union $\bigcup_{j=1}^v n_{ij}$, i.e., $P_{\text{s}}(n_i) \subsetneq \bigcup_{j=1}^v n_{ij}$. \(Prob(X)\) is the prediction probability of the model when the set of neurons \(X\) is activated, and \(T_{\text{low}}\) and \(T_{\text{high}}\) are predefined thresholds of acceptable prediction probability difference. Equation \eqref{equation-1} indicates that suppressing any proper subset of \(d^{\mathcal{D}}_i\), i.e., $P_{\text{s}}(n_i)$, will not result in a significant decrease in prediction probability; whereas Equation \eqref{equation-2} shows that suppressing all the neurons in \(d^{\mathcal{D}}_i\) will lead to a significant decrease in prediction probability. This demonstrates that these neurons store the same knowledge.

In the general case, considering that we have \(n\) knowledge neurons and we need to evaluate all possible subsets, the complexity of finding \(\mathcal{D}\) is \(O(2^n)\). To make the problem tractable, we simplified the problem by assuming that each \( d^{\mathcal{D}}_i \) only contains two knowledge neurons. This assumption reduces the problem complexity to \(O(n^2)\). 

To further reduce the computation, we design a two-step filtering process. Depicted in Algorithm \ref{alg:degenerate} and the module 3 of Figure \ref{fig:overall}, we first suppress each neuron and record neurons that do not cause a significant decrease in prediction probability, which are regarded as potential degenerate knowledge neurons $P_{\text{d}}$. For the elements in $P_{\text{d}}$, perform secondary filtering: suppress the pair of neurons in it, and if this operation leads to a significant decrease in the prediction probability of the model, record the pair of neurons as a degenerate knowledge neuron $d_i^{\mathcal{D}}$. Finally, we can return the degenerate knowledge neurons as \(\mathcal{D}\).
\begin{algorithm}[tb]
\caption{Identification of Degenerate Knowledge Neurons ($\mathcal{D}$)}
\label{alg:degenerate}
\textbf{Input}: Query $q$, thresholds $T_{\text{low}}$ and $T_{\text{high}}$.

\textbf{Output}: Degenerate knowledge neurons $\mathcal{D}$.

\begin{algorithmic}[1] 
\STATE Localize knowledge neurons \(\mathcal{N}=\{n_1, n_2, \ldots, n_k\}\).
\STATE Let \(P_{\text{d}} \leftarrow \emptyset\) (potential degenerate knowledge neurons).
\FOR{each \(n_i\) in \(\mathcal{N}\)}
\IF{\(Prob(\mathcal{N})\) - \(Prob(\mathcal{N} \setminus \{n_i\})\) \(\leq T_{\text{low}}\)}
\STATE \(P_{\text{d}} \leftarrow P_{\text{d}} \cup \{n_i\}\)
\ENDIF
\ENDFOR
\STATE Let \(\mathcal{D} \leftarrow \emptyset\) (degenerate knowledge neurons).
\FOR{each \(n_{i1}, n_{i2}\) in \(P_{\text{d}}\), \(n_{i1} \neq n_{i2}\)}
\IF{$Prob(\mathcal{N})$ - \(Prob(\mathcal{N} \setminus \{n_{i1}, n_{i2}\})\) \(> T_{\text{high}}\)}
\STATE \(\mathcal{D} \leftarrow \mathcal{D} \cup \{n_{i1}, n_{i2}\}\),
where $\{n_{i1}, n_{i2}\}$ is a $d_{i}^{\mathcal{D}}$ within \(\mathcal{D}\).
\ENDIF
\ENDFOR
\STATE \textbf{return} \(\mathcal{D}\)
\end{algorithmic}
\end{algorithm}

\section{Experiments}
\subsection{Experimental Settings}
\vpara{Model Selection and Dataset}
For our experiments, we opt for two distinct multilingual PLMs: m-BERT~\cite{mbert} and m-GPT~\cite{mgpt}. The m-BERT, an auto-encoding model, is pre-trained on a diverse collection of multilingual data, while the m-GPT, an auto-regressive model, is designed to process a wide-ranging corpus of 61 languages. Regarding the datasets, we employ mLAMA \cite{mlama}, a multilingual extension of the original LAMA~\cite{fill-in-the-blank,lama2} to localize the knowledge in multilingual PLMs. 

\vpara{Evaluation Metrics}
We apply the same neuron editing manipulation to both methods, where the detected knowledge neurons are suppressed or enhanced, followed by calculating the prediction probability of the PLM for both relevant and irrelevant facts. To compare the precision of knowledge localization across different methods in a comprehensive manner, we propose a new evaluation metric to assess the results of knowledge localization across the entire dataset:
{\small\begin{align}
    SR_x = \frac{{\Delta Prob_{rx}}}{{\Delta Prob_{ix}}}
\end{align}
}where \( SR_x \) is the editing success rate, and \( x \) represents our editing operation to suppress or enhance the neurons. Given a query, it itself is considered as a relevant fact, and a fact of a different type is randomly selected as its irrelevant fact. $\Delta Prob_{rx}$ and $\Delta Prob_{ix}$ represent the average changes in prediction probability under operation $x$ for relevant and irrelevant facts, respectively. Overall, we hope that relevant facts change with the change of knowledge neurons, while irrelevant facts remain unchanged; thus, the higher the success rate, the better the localization results\footnote{Data with extremely high $\Delta Prob_{rx}$ or \( \Delta Prob_{ix} \), reflecting unmastered facts, is excluded in order to localize storage of mastered facts.}. Since we perform suppress and enhancement operations on neurons separately, the success rates of these two cases are summed up as the final success rate: $SR = {SR_{\text{enhance}}}+{SR_{\text{suppress}}}$.

\begin{figure}
   \centering
   \subfloat{
     \includegraphics[width=0.8\linewidth]{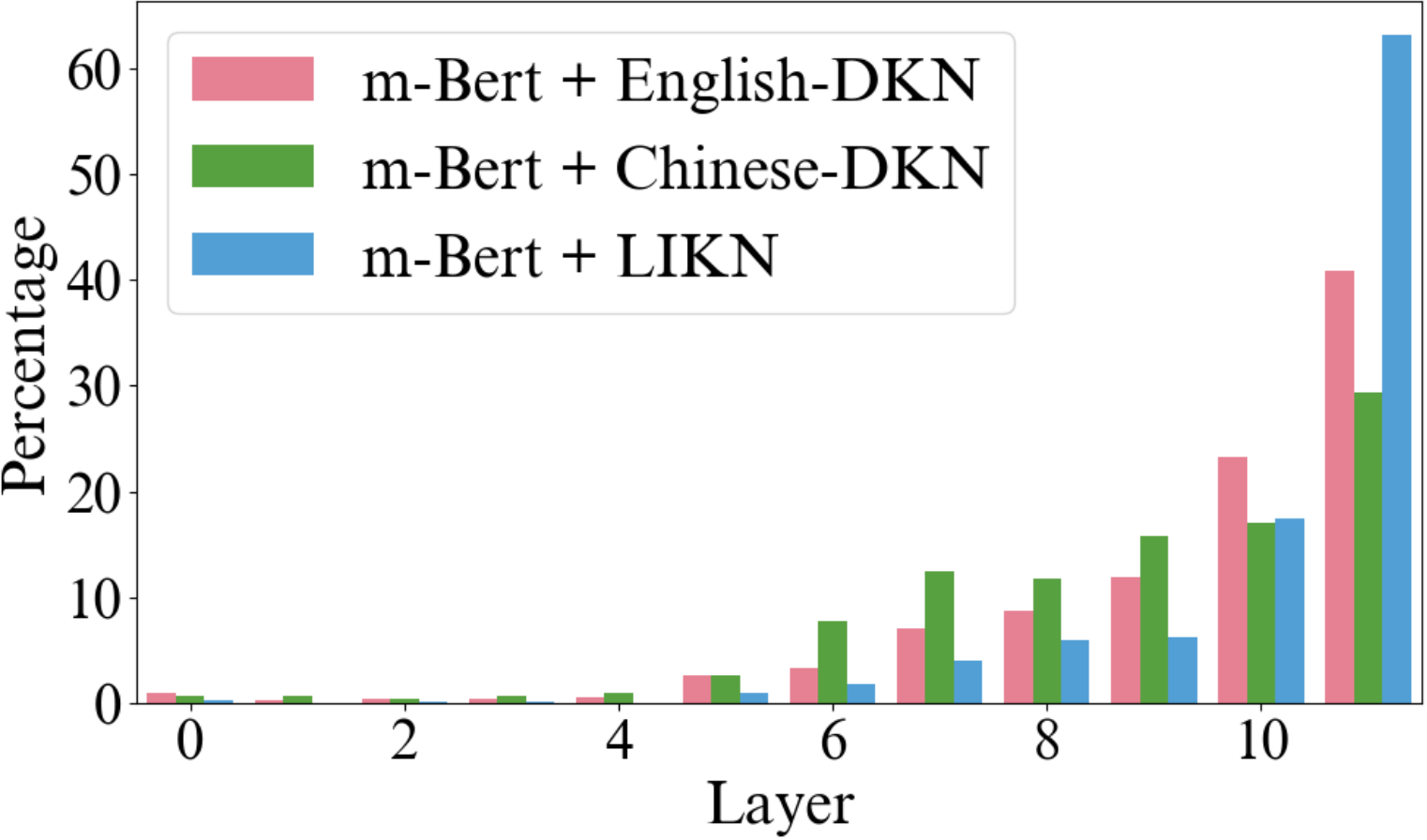}
     }
   \\
   \subfloat{
     \includegraphics[width=0.8\linewidth]{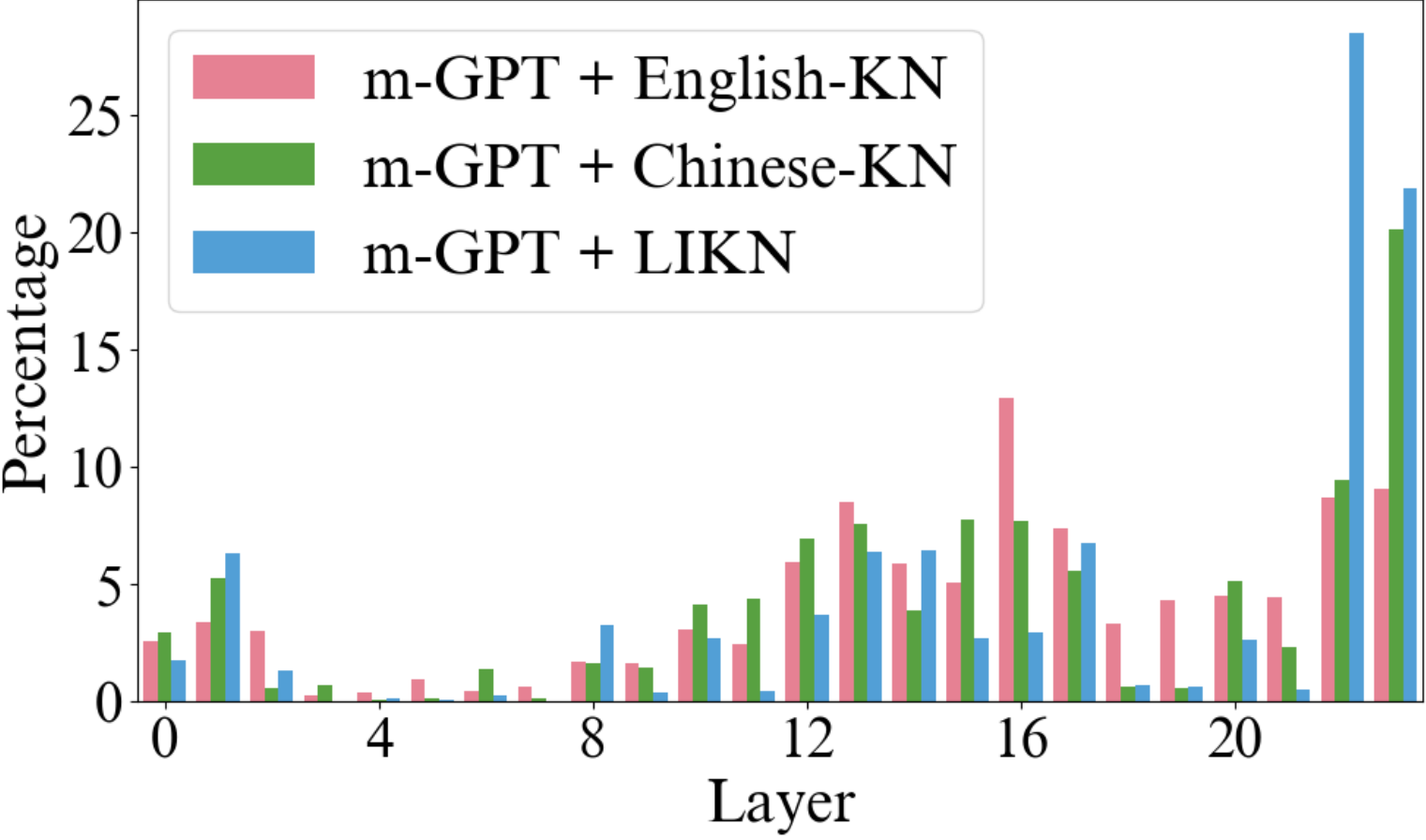}
   }
   \caption{The distributions of knowledge neurons in m-BERT and m-GPT models under two languages (English-KN and Chinese-KN) and language-independent knowledge neurons (LIKN).}
   \label{fig3} 
\end{figure}

\subsection{Localization of Knowledge Neurons}
We carry out experiments using the module 1 on both m-BERT and m-GPT models across English and Chinese datasets, and take the method proposed by \citeauthor{dai2022kn}\shortcite{dai2022kn} as the baseline, which we denote as B-KN.
The findings from our study are presented in Table \ref{table1} and Figure \ref{fig3}, from which we derive several key insights.

(1) Our method achieves better results in all settings. In Table \ref{table1}, we use AMIG to represent our method, and the results in the table represents the average success rate \( SR \). Under all settings, our method outperforms B-KN, especially for the Chinese dataset, where the success rates for m-BERT and m-GPT have increased by 84.34\% and 44.49\% respectively. This demonstrates that the knowledge neurons localized by our method are more precise.

(2) In m-BERT, knowledge neurons are primarily in the final layers, whereas in m-GPT they are in the early, middle, and final layers, as shown in Figure \ref{fig3}, where the \textit{x} and \textit{y} axes represent the PLM layers and the percentage of knowledge neurons, respectively. This might be due to the auto-encoding models (e.g., m-BERT), which share encoding space and encode high-level features in the final few layers, while the auto-regressive models (e.g., m-GPT) gradually refine the features at each layer to predict the next word.

(3) The distributions of knowledge neurons for Chinese and English are relatively similar, but differences persist. Similarities could be due to facts having the same meaning across languages, while differences might result from the inherent structural and syntactic differences between the languages or from variations in the quality of the pretraining corpora.
\begin{table}
\centering
\scalebox{0.84}{
\begin{tabular}{l|l|cc}
\midrule
Dataset & Method & m-BERT & m-GPT \\
\midrule
\multirow{2}{*}{English} & B-KN & 3.944 & 5.207 \\
                         & AMIG (Ours) & \textbf{4.035} (\textuparrow\,2.31\%) & \textbf{5.603} (\textuparrow\,7.61\%) \\
\midrule
\multirow{2}{*}{Chinese} & B-KN & 5.579 & 5.439 \\
                         & AMIG (Ours) & \textbf{10.285} (\textuparrow\,84.34\%) & \textbf{7.859} (\textuparrow\,44.49\%) \\
\midrule
\end{tabular}}
\caption{Results of the localization of knowledge neurons. B-KN is the baseline method, the symbol ``\textuparrow" indicates the increase in success rate compared to B-KN for our method, which can be expressed as: $\frac{{\text{AMIG}-\text{B-KN}}}{{\text{B-KN}}}$, and bold indicates the method with a higher $SR$.}
\label{table1}
\end{table}

\subsection{Language-Independence Neurons and Cross-Lingual Knowledge Editing}
\vpara{Localization of Language-Independence Neurons} 
Through our experiment with module 2, we capture the results in Figure \ref{fig3}. The findings reveal that, whether in m-BERT or m-GPT, language-independent knowledge neurons are primarily concentrated in the final one or two layers. 
This might be because language-independent facts serve as high-level features, and the PLM is only able to successfully encode them in the final few layers.

\vpara{Cross-Lingual Knowledge Editing Experimental Settings and Results}
We design cross-lingual editing experiments based on language-independent knowledge neurons. Similar to the setup of knowledge localization experiments, we suppress or enhance language-independent knowledge neurons and calculate the editing success rate \( SR \). To demonstrate the role of language-independent knowledge neurons, we design two comparative experiments:
(1) Editing the knowledge neurons of one language and observing the changes in the corresponding facts in another language. 
(2) Sequentially editing the knowledge neurons of two languages, observing the changes in the corresponding facts in both languages.

Our analysis of Table \ref{table:cross-lingual-edit} brings to light two insights.

(1) Language-independent knowledge neurons facilitate cross-lingual editing. Compared to editing in Chinese or English, editing language-independent knowledge neurons has a higher success rate in all settings; in the Chinese dataset, the success rates for m-BERT and m-GPT increased by 213.05\% and 277.36\%. This indicates that while editing facts in one language and expecting changes in another is challenging, language-independent neurons provide a viable solution.

(2) Editing each language separately does not guarantee better results. Though one might intuitively edit each language to achieve cross-lingual changes, our experiments show that this method not only relies on more computational resources but also might underperform. Sequential editing led to 42.97\% and 58.80\% lower success rates for m-BERT and m-GPT respectively, compared to using language-independent neurons, possibly due to confusion from multiple edits. This emphasizes the importance of language-independent neurons.
\begin{table}
\centering
\scalebox{0.82}{
\begin{tabular}{l|l|cc}
\toprule
Dataset & Method & m-BERT & m-GPT \\
\midrule
\multirow{3}{*}{English} & LIKN (Ours) & 2.359 (\textuparrow\,10.29\%) & 2.542 (\textuparrow\,5.92\%) \\
                         & Mono-KN & 2.139 & 2.400 \\
                         & Seq-KN & 3.800 & 4.285 \\
\midrule
\multirow{3}{*}{Chinese} & LIKN (Ours) & 7.175 (\textuparrow\,213.05\%) & 8.868 (\textuparrow\,277.36\%) \\
                         & Mono-KN & 2.292 & 2.350 \\
                         & Seq-KN & 4.092 (\textdownarrow\,42.97\%) & 3.654 (\textdownarrow\,58.80\%) \\
\bottomrule
\end{tabular}}
\caption{Results of cross-lingual knowledge editing. LIKN represents editing language-independent knowledge neurons, Mono-KN denotes editing knowledge neurons in one language's dataset corresponding to another, and Seq-KN denotes sequentially editing knowledge neurons in two languages. The symbol `\textuparrow` shows a success rate increase in LIKN over Mono-KN, represented as $\frac{\text{LIKN} - \text{Mono-KN}}{\text{Mono-KN}}$, and `\textdownarrow` indicates a decrease in LIKN compared to Seq-KN, represented as $\frac{\text{LIKN} - \text{Seq-KN}}{\text{LIKN}}$.}

\label{table:cross-lingual-edit}
\end{table}
\subsection{Degenerate Knowledge Neurons and Fact-Checking Experiment}
\begin{figure}[h]
  \centering
  \subfloat{
    \includegraphics[width=0.8\linewidth]{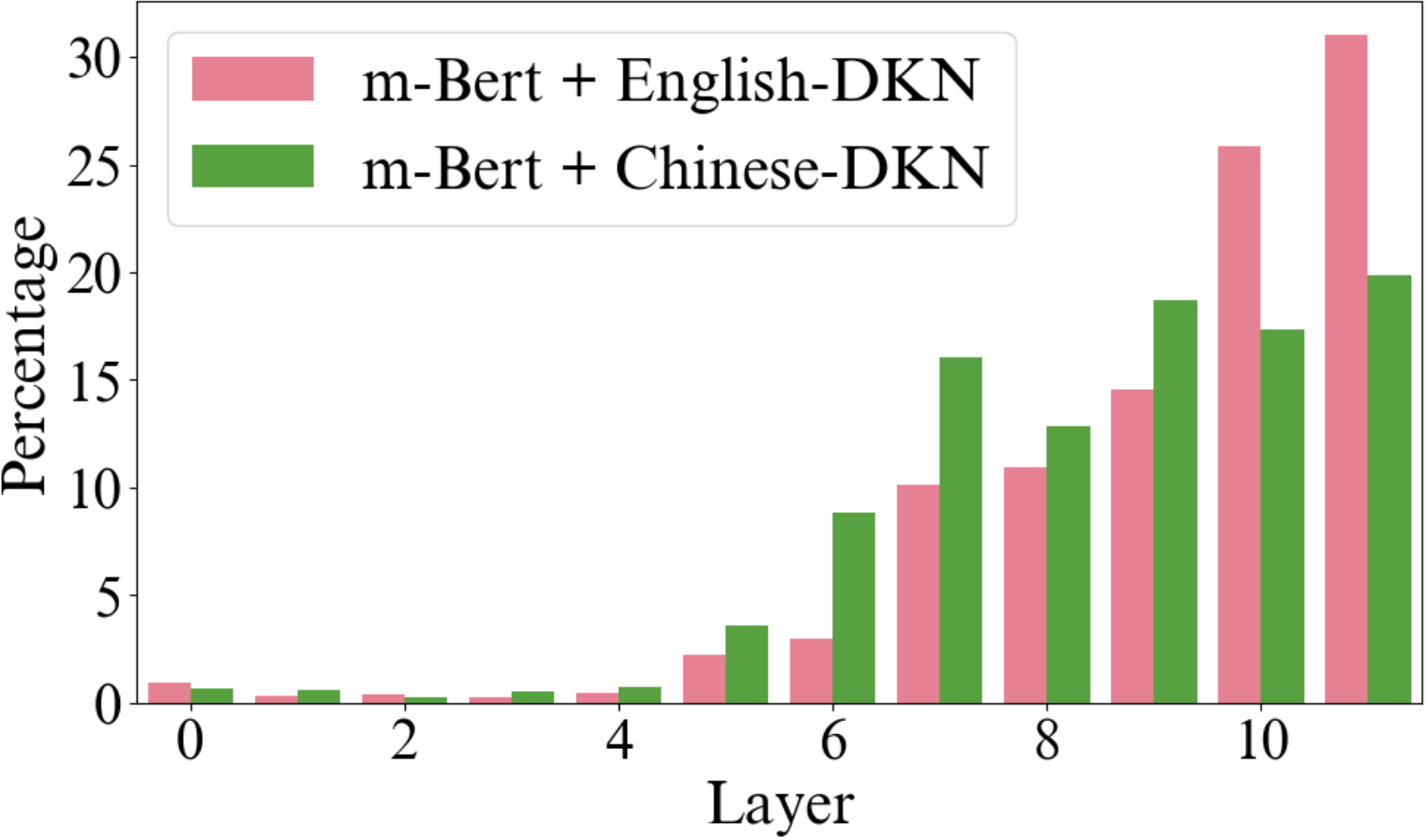}
    \label{fig:d-mbert}
  }
  \\
  \subfloat{
    \includegraphics[width=0.8\linewidth]{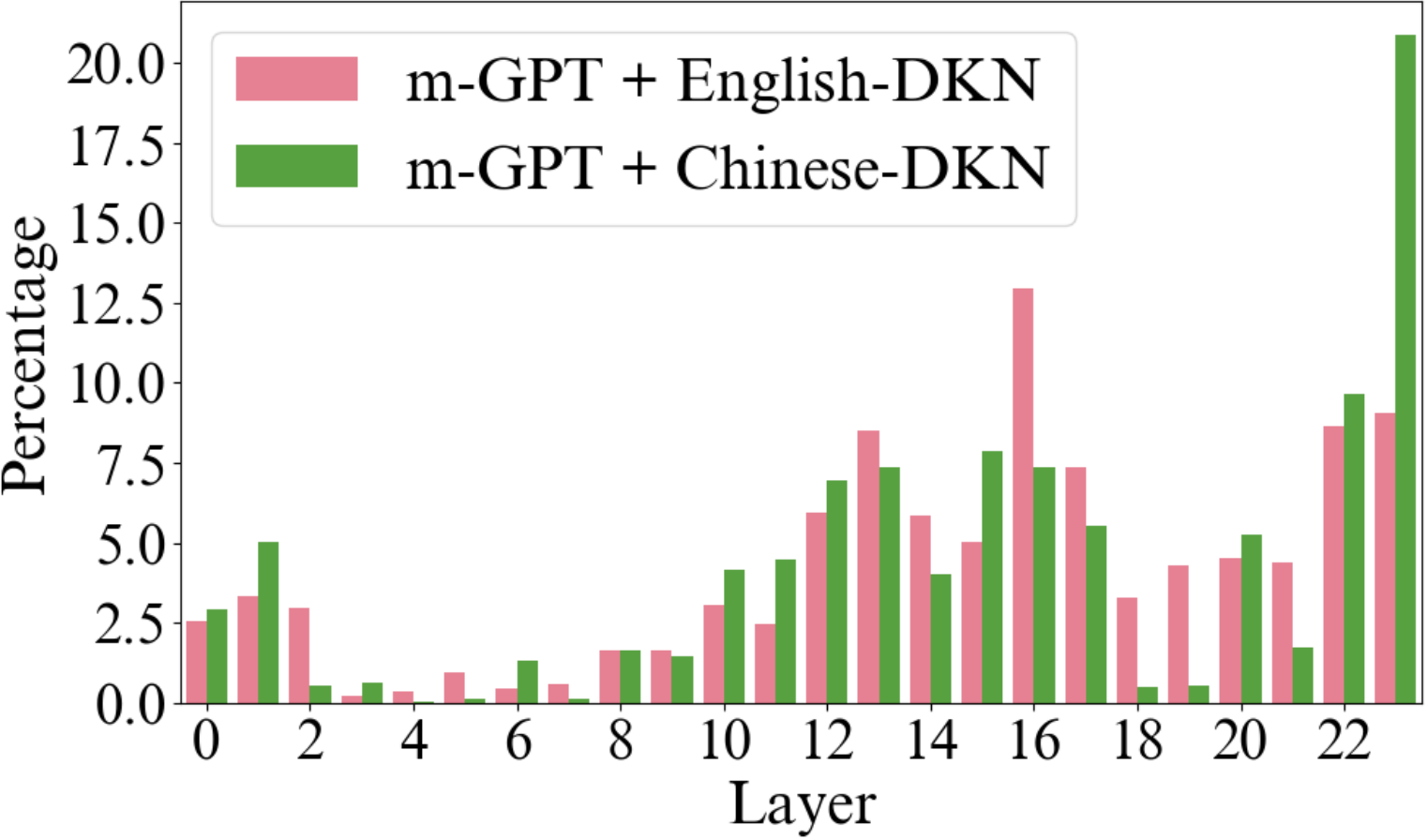}
    \label{fig:d-mgpt}
  }
  \caption{The distributions of degenerate knowledge neurons (DKN) in multilingual PLMs under two languages.}
  \label{fig:d}
\end{figure}

\vpara{Identification of Degenerate Knowledge Neurons in Multilingual PLMs} 
We set up an experiment using module 3 to investigate the degenerate knowledge neurons, and the results are displayed in Figure \ref{fig:d}. From our observations, degenerate knowledge neurons in m-BERT and m-GPT exhibit distribution patterns similar to knowledge neurons. This not only demonstrates a strong correlation between the degeneracy of factual knowledge and the facts themselves, but also reflects the PLMs' mastery of the facts.

\begin{figure}[h]
  \centering
  \includegraphics[width=0.8\linewidth]{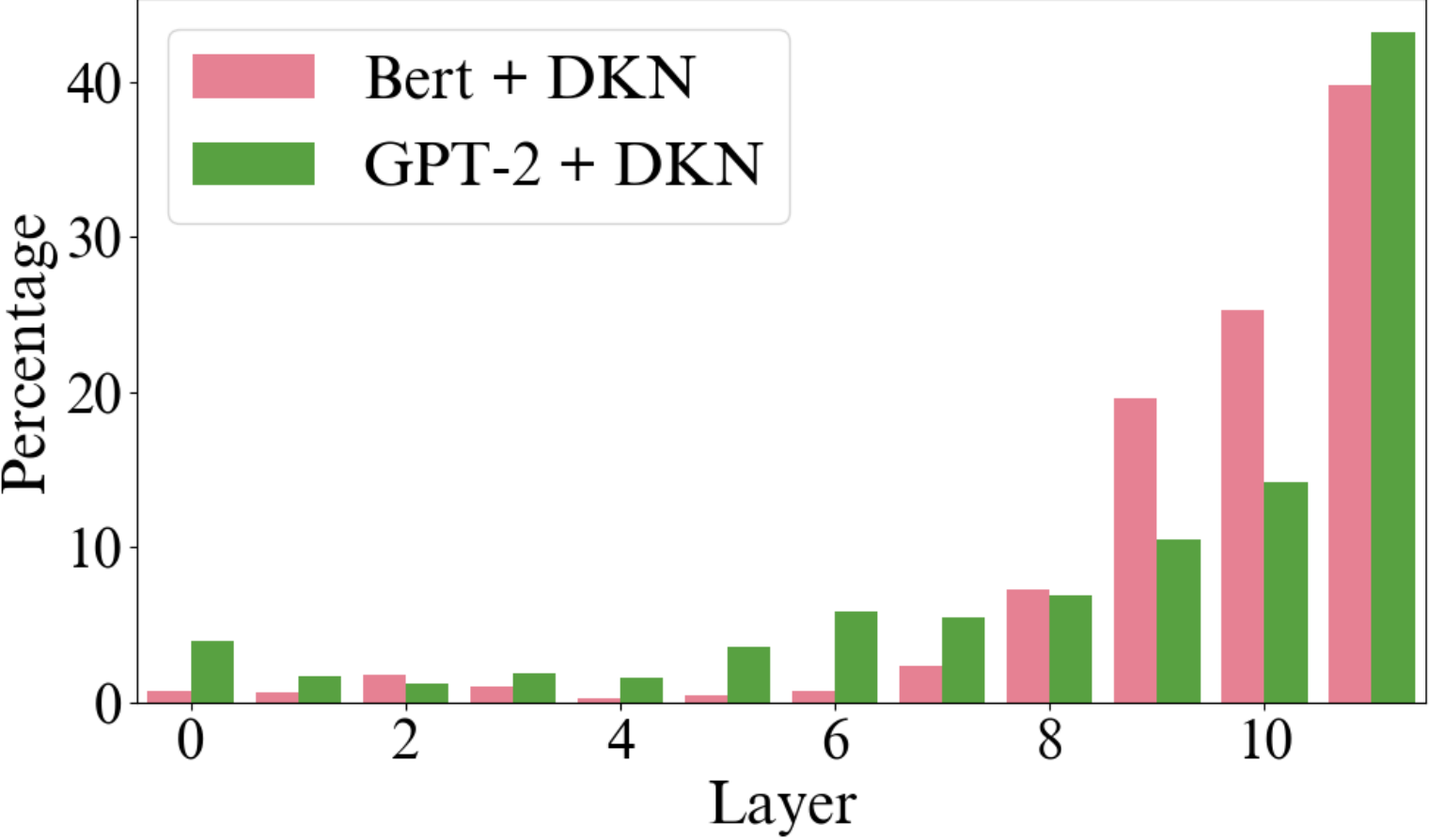}
  \caption{The distributions of degenerate knowledge neurons (DKN) in monolingual PLMs under two languages.}
  \label{fig:mono-D-neurons_distribution}
\end{figure}
\vpara{Identification of degenerate knowledge neurons in Monolingual PLMs}
In our experiments with monolingual PLMs, we successfully identify the degenerate knowledge neurons and prove that they are inherently present within the PLMs. 
A possible question regarding degenerate knowledge neurons is: does the PLMs store the same fact in multiple languages, thus utilizing multiple neuron sets for the same information? To dispel this notion and demonstrate that the existence of degenerate knowledge neurons is unrelated to the support of multilingualism in the PLMs, we extend our exploration to monolingual PLMs, specifically in BERT and GPT-2. The distributions of these degenerate knowledge neurons is depicted in Figure \ref{fig:mono-D-neurons_distribution}, further reinforcing our conclusion.

\vpara{Fact-Checking Experimental Settings and Results}
PLMs may conceal false facts \cite{hallucination_chatgpt1,hallucination_chatgpt2}, and current solutions often rely on external data for fact-checking \cite{fact-checking-survey}. 
Considering the nature of the functional overlap of degenerate knowledge neurons, we design a fact-checking experiment to detect wrong facts based on degenerate knowledge neurons without relying on external data. Next, we introduce our experimental settings in detail.

 First, the mLAMA dataset is modified to include a \textit{wrong fact} attribute.
 For a fact triple associated with a certain relation name of fact, such as $\langle \textit{Tanzania}, \textit{Capital}, \textit{Dodoma} \rangle$ , we randomly select an object (e.g., \textit{Dar es Salaam}) from the same relation name as a wrong fact. 
Then, to validate the practical implications of our findings, we divide each type of query in the dataset into two parts proportionally. For each type, the first segment is used to obtain degenerate knowledge neurons, and we identify those exceeding a certain threshold of \( t\% \) in quantity.
Then, we take the queries from the second part, along with the corresponding correct or incorrect facts, as input and compute the average activation score of the degenerate knowledge neurons. If the average activation score surpasses a threshold $\lambda$, the fact is classified as correct.
We use the original PLMs to directly evaluate the correctness of facts for comparative analysis. This configuration prevents the PLMs from employing the degenerate knowledge neurons of the query itself for fact-checking, rendering the experiments more convincing. We denote our method as ``with\_DKN" in the Table \ref{table:combined-experiments}.
Finally, since the current fact-checking method must rely on external data, we use the PLMs to directly perform fact-checking as the baseline of our method, denoted as ``wo\_DKN" in the Table \ref{table:combined-experiments}. We use Precision, Recall and F1-score as evaluation metrics.

The results in Table \ref{table:combined-experiments} lead us to the following conclusions.

(1) Degenerate knowledge neurons can help the PLMs detect wrong facts. Under various settings, our method is better than the baseline method, especially for Chinese datasets and auto-regressive models. For instance, in the context of m-GPT and Chinese datasets, the F1 score of our method has increased by 167150\% compared to the baseline.This substantial improvement indicates that the presence of degenerate knowledge neurons enhances the PLMs' stable mastery of factual knowledge. 

(2) Using PLMs for fact-checking, they often judge a fact as correct, leading to extremely high Recall. This aligns with observations that generative language models may produce incorrect information if presented with a false premise \cite{hallucination_chatgpt3, lakshmanan2022large, metz2022new}.
It is essential to recognize that a model's low predictive probability does not hinder the accurate identification of knowledge neurons. As shown in Equation~\ref{eq:1}, using the true value \( y^* \) allows for correct knowledge neuron localization even when the model's output is erroneous.

(3) Auto-regressive models show higher Recall than auto-encoding models. This may be due to the auto-regressive design favoring coherence over accuracy, and the auto-encoding possibly being more conservative  \cite{zhou2023comprehensive}.

(4) The existence of degenerate knowledge neurons is unrelated to the support of multilingualism in the PLMs. In the monolingual PLMs, i.e., BERT and GPT-2, fact-checking can also be performed based on degenerate knowledge neurons. This result further proves the existence of degenerate knowledge neurons and its usefulness.
\begin{table}
\centering
\scalebox{0.69}{
\begin{tabular}{l|l|c|ccc}
\midrule
Dataset & Model & Method & P & R & F1 \\
\midrule
\multirow{4}{*}{English} & \multirow{2}{*}{m-BERT} & wo\_DKN & 0.222 & 0.986 & 0.362 \\
                          &                        & with\_DKN (Ours) & 0.493 & 0.599 & \textbf{0.541} (\textuparrow\,49\%) \\
\cmidrule{2-6}
                          & \multirow{2}{*}{m-GPT}  & wo\_DKN & 0.010 & 1.000 & 0.021 \\
                          &                        & with\_DKN (Ours) & 0.311 & 0.709 & \textbf{0.433} (\textuparrow\,1962\%) \\
\midrule\midrule
\multirow{4}{*}{Chinese} & \multirow{2}{*}{m-BERT} & wo\_DKN & 0.010 & 1.000 & 0.020 \\
                          &                        & with\_DKN (Ours) & 0.870 & 0.524 & \textbf{0.654} (\textuparrow\,3170\%) \\
\cmidrule{2-6}
                          & \multirow{2}{*}{m-GPT}  & wo\_DKN & 0.0002 & 1.000 & 0.0004 \\
                          &                        & with\_DKN (Ours) & 0.966 & 0.511 & \textbf{0.669} (\textuparrow\,167150\%) \\
\midrule\midrule
\multirow{4}{*}{English} & \multirow{2}{*}{BERT}   & wo\_DKN & 0.301 & 0.983 & 0.460 \\
                          &                        & with\_DKN (Ours) & 0.504 & 0.571 & \textbf{0.535} (\textuparrow\,16\%) \\
\cmidrule{2-6}
                          & \multirow{2}{*}{GPT-2}  & wo\_DKN & 0.010 & 1.000 & 0.021 \\
                          &                        & with\_DKN (Ours) & 0.315 & 0.608 & \textbf{0.415} (\textuparrow\,1876\%) \\
\midrule
\end{tabular}
}
\caption{Fact-checking experiment results comparing methods with (with\_DKN) and without (wo\_DKN) degenerate knowledge neurons. The symbol ``\textuparrow" shows F1-score improvement in with\_DKN over wo\_DKN as $\frac{\text{with\_DKN} - \text{wo\_DKN}}{\text{wo\_DKN}}$, with bold indicating the higher score.}
\label{table:combined-experiments}
\end{table}

\section{Related Work}
\vpara{Knowledge Localization}
Existing methods roughly fall into two categories: (1) Gradient-based method: \citeauthor{dai2022kn}\shortcite{dai2022kn} first introduces the concept of knowledge neurons and localizes them by assessing the contribution of each neuron \cite{key_value} through calculating their attribution scores using integrated gradients. (2) Causal-inspired method, introduced by \citeauthor{meng2022locating}\shortcite{meng2022locating}, defines knowledge neurons as the neuron activations within PLMs that have the strongest causal effect on predicting certain factual knowledge, and this method has inspired the creation of knowledge editing algorithms such as ROME \cite{meng2022locating}, MEMIT \cite{meng2022memit}, and MEND \cite{mend}. However, current methods lack a universal approach for different PLM architectures and exploration in multiple languages.

\vpara{Axiomatic Attribution Methods}
\citeauthor{ig}\shortcite{ig} introduces the axiomatic attribution method, emphasizing Sensitivity and Implementation Invariance as the core axioms for attribution methods, leading to Integrated Gradients (IG). Subsequent research includes Discretized IG~\cite{DIG}, which uses interpolation strategies for gradient accuracy; Sequential IG~\cite{enguehard2023sequential} designed for word importance evaluation; and Effective Shapley value along with Shapley IG, developed by \citeauthor{transparent}\shortcite{transparent} to enhance efficiency and effectiveness. We improve the baseline vectors for IG to minimize their information content.

\section{Conclusion}
In this research, we explore factual knowledge localization in multilingual PLMs using our architecture-adapted multilingual integrated gradient method. We further design two modules, leading to two discoveries of language-independent knowledge neurons and degenerate knowledge neurons. The former affirms that a portion of the knowledge in multilingual PLMs exists in a form that transcends language, while the latter introduces a novel type of neuron which is similar to the degeneration phenomenon observed in biological systems, and these neurons can be used to detect incorrect facts.

\section*{Acknowledgments}
This work is supported by the National Key Research and Development Program of China (No. 2020AAA0106400), the National Natural Science Foundation of China (No. 61976211, 62176257). This work is also supported by the Strategic Priority Research Program of Chinese Academy of Sciences (Grant No.XDA27020100), the Youth Innovation Promotion Association CAS, and Yunnan Provincial Major Science and Technology Special Plan Projects (No.202202AD080004).

\bibliography{aaai24.bib}

\end{document}